\documentclass{article}
\usepackage{ijcai17}

\usepackage{times}

\usepackage{amsfonts}
\usepackage{amsmath}
\usepackage{graphicx}
\usepackage{caption}
\usepackage{subcaption}
\usepackage{algorithm}
\usepackage{algpseudocode}
\usepackage{multirow}
\usepackage{array}

\title{A Generalized Recurrent Neural Architecture \\ for Text Classification with Multi-Task Learning}
\author{Honglun Zhang$^{1}$, Liqiang Xiao$^{1}$, Yongkun Wang$^{2}$, Yaohui Jin$^{1,2}$ \\
$^{1}$State Key Lab of Advanced Optical Communication System and Network \\
$^{2}$Network and Information Center \\ 
Shanghai Jiao Tong University \\
\{ykw\}@sjtu.edu.cn}

\begin{document}

\maketitle

\begin{abstract}
  
Multi-task learning leverages potential correlations among related tasks to extract common features and yield performance gains. However, most previous works only consider simple or weak interactions, thereby failing to model complex correlations among three or more tasks. In this paper, we propose a multi-task learning architecture with four types of recurrent neural layers to fuse information across multiple related tasks. The architecture is structurally flexible and considers various interactions among tasks, which can be regarded as a generalized case of many previous works. Extensive experiments on five benchmark datasets for text classification show that our model can significantly improve performances of related tasks with additional information from others.

\end{abstract}

\section{Introduction}

Neural network based models have been widely exploited with the prosperities of \textit{Deep Learning}~\cite{DBLP:journals/pami/BengioCV13} and achieved inspiring performances on many NLP tasks, such as text classification~\cite{DBLP:conf/emnlp/ChenQZWH15,DBLP:conf/emnlp/LiuQCWH15}, semantic matching~\cite{DBLP:conf/emnlp/LiuQZCH16,DBLP:conf/acl/LiuQCH16} and machine translation~\cite{DBLP:conf/nips/SutskeverVL14}. These models are robust at feature engineering and can represent words, sentences and documents as fix-length vectors, which contain rich semantic information and are ideal for subsequent NLP tasks. 

One formidable constraint of deep neural networks (DNN) is their strong reliance on large amounts of annotated corpus due to substantial parameters to train. A DNN trained on limited data is prone to overfitting and incapable to generalize well. However, constructions of large-scale high-quality labeled datasets are extremely labor-intensive. To solve the problem, these models usually employ a pre-trained lookup table, also known as \textit{Word Embedding}~\cite{DBLP:conf/nips/MikolovSCCD13}, to map words into vectors with semantic implications. However, this method just introduces extra knowledge and does not directly optimize the targeted task. The problem of insufficient annotated resources is not solved either.

Multi-task learning leverages potential correlations among related tasks to extract common features, increase corpus size implicitly and yield classification improvements. Inspired by ~\cite{DBLP:journals/ml/Caruana97}, there are a large literature dedicated for multi-task learning with neural network based models~\cite{DBLP:conf/icml/CollobertW08,DBLP:conf/naacl/LiuGHDDW15,DBLP:conf/emnlp/LiuQH16,DBLP:conf/ijcai/LiuQH16}. These models basically share some lower layers to capture common features and further feed them to subsequent task-specific layers, which can be classified into three types: 

\begin{itemize}
\item \textbf{Type-\uppercase\expandafter{\romannumeral1}} One dataset annotated with multiple labels and one input with multiple outputs.
\item \textbf{Type-\uppercase\expandafter{\romannumeral2}} Multiple datasets with respective labels and one input with multiple outputs, where samples from different tasks are fed one by one into the models sequentially. 
\item \textbf{Type-\uppercase\expandafter{\romannumeral3}} Multiple datasets with respective labels and multiple inputs with multiple outputs, where samples from different tasks are jointly learned in parallel.
\end{itemize}

In this paper, we propose a generalized multi-task learning architecture with four types of recurrent neural layers for text classification. The architecture focuses on \textbf{Type-\uppercase\expandafter{\romannumeral3}}, which involves more complicated interactions but has not been researched yet. All the related tasks are jointly integrated into a single system and samples from different tasks are trained in parallel. In our model, every two tasks can directly interact with each other and selectively absorb useful information, or communicate indirectly via a shared intermediate layer. We also design a global memory storage to share common features and collect interactions among all tasks.

We conduct extensive experiments on five benchmark datasets for text classification. Compared to learning separately, jointly learning multiple relative tasks in our model demonstrate significant performance gains for each task. 

Our contributions are three-folds:

\begin{itemize}
\item Our model is structurally flexible and considers various interactions, which can be concluded as a generalized case of many previous works with deliberate designs.
\item Our model allows for interactions among three or more tasks simultaneously and samples from different tasks are trained in parallel with multiple inputs.
\item We consider different scenarios of multi-task learning and demonstrate strong results on several benchmark classification datasets. Our model outperforms most of state-of-the-art baselines.
\end{itemize}

\section{Problem Statements}

\subsection{Single-Task Learning}

For a single supervised text classification task, the input is a word sequences denoted by $x=\{x_1,x_2,...,x_T\}$, and the output is the corresponding class label $y$ or class distribution $\mathbf{y}$. A lookup layer is used first to get the vector representation $\mathbf{x}_i\in\mathbb{R}^d$ of each word $x_i$. A classification model $f$ is trained to transform each $\mathbf{x}=\{\mathbf{x}_1,\mathbf{x}_2,...,\mathbf{x}_T\}$ into a predicted distribution $\hat{\mathbf{y}}$.
\begin{equation}\tag{$1$}\label{eq:1}
f(\mathbf{x}_1,\mathbf{x}_2,...,\mathbf{x}_T)=\hat{\mathbf{y}}
\end{equation}
and the training objective is to minimize the total cross-entropy of the predicted and true distributions over all samples.
\begin{equation}\tag{$2$}\label{eq:2}
L=-\sum_{i=1}^{N}\sum_{j=1}^{C}y_{ij}\log{\hat{y}_{ij}}
\end{equation}
where $N$ denotes the number of training samples and $C$ is the class number.

\subsection{Multi-Task Learning}

Given $K$ supervised text classification tasks, $T_1,T_2,...,T_K$, a jointly learning model $\mathbf{F}$ is trained to transform multiple inputs into a combination of predicted distributions in parallel.
\begin{equation}\tag{$3$}\label{eq:3}
\mathbf{F}(\mathbf{x}^{(1)}, \mathbf{x}^{(2)},...,\mathbf{x}^{(K)})=(\hat{\mathbf{y}}^{(1)},\hat{\mathbf{y}}^{(2)},...,\hat{\mathbf{y}}^{(K)})
\end{equation}
where $\mathbf{x}^{(k)}$ are sequences from each tasks and $\hat{\mathbf{y}}^{(k)}$ are the corresponding predictions.

The overall training objective of $\mathbf{F}$ is to minimize the weighted linear combination of costs for all tasks.
\begin{equation}\tag{$4$}\label{eq:4}
\mathbf{L}=-\sum_{i=1}^{N}\sum_{k=1}^{K}\lambda_k\sum_{j=1}^{C_k}y_{ij}^{(k)}\log{\hat{y}_{ij}^{(k)}}
\end{equation}
where $N$ denotes the number of sample collections, $C_k$ and $\lambda_k$ are class numbers and weights for each task $T_k$ respectively.

\subsection{Three Perspectives of Multi-Task Learning}\label{sec:scenarios}

Different tasks may differ in characteristics of the word sequences $\mathbf{x}$ or the labels $\mathbf{y}$. We compare lots of benchmark tasks for text classification and conclude three different perspectives of multi-task learning.

\begin{itemize}
\item \textbf{Multi-Cardinality} Tasks are similar except for cardinality parameters, for example, movie review datasets with different average sequence lengths and class numbers. 
\item \textbf{Multi-Domain} Tasks involve contents of different domains, for example, product review datasets on books, DVDs, electronics and kitchen appliances. 
\item \textbf{Multi-Objective} Tasks are designed for different objectives, for example, sentiment analysis, topics classification and question type judgment.
\end{itemize}

The simplest multi-task learning scenario is that all tasks share the same cardinality, domain and objective, while come from different sources, so it is intuitive that they can obtain useful information from each other. However, in the most complex scenario, tasks may vary in cardinality, domain and even objective, where the interactions among different tasks can be quite complicated and implicit. We will evaluate our model on different scenarios in the Experiment section.

\section{Methodology}

Recently neural network based models have obtained substantial interests in many natural language processing tasks for their capabilities to represent variable-length text sequences as fix-length vectors, for example, Neural Bag-of-Words (NBOW), Recurrent Neural Networks (RNN), Recursive Neural Networks (RecNN) and Convolutional Neural Network (CNN). Most of them first map sequences of words, n-grams or other semantic units into embedding representations with a pre-trained lookup table, then fuse these vectors with different architectures of neural networks, and finally utilize a softmax layer to predict categorical distribution for specific classification tasks. For recurrent neural network, input vectors are absorbed one by one in a recurrent way, which makes RNN particularly suitable for natural language processing tasks.

\subsection{Recurrent Neural Network}

A recurrent neural network maintains a internal hidden state vector $\mathbf{h}_t$ that is recurrently updated by a transition function $f$. At each time step $t$, the hidden state $\mathbf{h}_t$ is updated according to the current input vector $\mathbf{x}_t$ and the previous hidden state $\mathbf{h}_{t-1}$.
\begin{equation}\tag{$5$}\label{eq:5}
\mathbf{h}_t=\left\{
\begin{array}{ll}
0 & t=0\\
f(\mathbf{h}_{t-1},\mathbf{x}_t) & \mbox{otherwise}
\end{array}
\right.
\end{equation}
where $f$ is usually a composition of an element-wise nonlinearity with an affine transformation of both $\mathbf{x}_t$ and $\mathbf{h}_{t-1}$. 

In this way, recurrent neural networks can comprehend a sequence of arbitrary length into a fix-length vector and feed it to a softmax layer for text classification or other NLP tasks. However, gradient vector of $f$ can grow or decay exponentially over long sequences during training, also known as the \textit{gradient exploding} or \textit{vanishing} problems, which makes it difficult to learn long-term dependencies and correlations for RNNs.

~\cite{DBLP:journals/neco/HochreiterS97} proposed Long Short-Term Memory Network (LSTM) to tackle the above problems. Apart from the internal hidden state $\mathbf{h}_t$, LSTM also maintains a internal hidden memory cell and three gating mechanisms. While there are numerous variants of the standard LSTM, here we follow the implementation of ~\cite{DBLP:journals/corr/Graves13}. At each time step $t$, states of the LSTM can be fully represented by five vectors in $\mathbb{R}^n$, an \textit{input gate} $\mathbf{i}_t$, a \textit{forget gate} $\mathbf{f}_t$, an \textit{output gate} $\mathbf{o}_t$, the \textit{hidden state} $\mathbf{h}_t$ and the \textit{memory cell} $\mathbf{c}_t$, which adhere to the following transition functions.
\begin{align*}
&\mathbf{i}_t=\sigma(\mathbf{W}_i\mathbf{x}_t+\mathbf{U}_i\mathbf{h}_{t-1}+\mathbf{V}_i\mathbf{c}_{t-1}+\mathbf{b}_i)\tag{$6$}\label{eq:6}\\
&\mathbf{f}_t=\sigma(\mathbf{W}_f\mathbf{x}_t+\mathbf{U}_f\mathbf{h}_{t-1}+\mathbf{V}_f\mathbf{c}_{t-1}+\mathbf{b}_f)\tag{$7$}\label{eq:7}\\
&\mathbf{o}_t=\sigma(\mathbf{W}_o\mathbf{x}_t+\mathbf{U}_o\mathbf{h}_{t-1}+\mathbf{V}_o\mathbf{c}_{t-1}+\mathbf{b}_o)\tag{$8$}\label{eq:8}\\
&\tilde{\mathbf{c}}_t=\tanh(\mathbf{W}_c\mathbf{x}_t+\mathbf{U}_c\mathbf{h}_{t-1})\tag{$9$}\label{eq:9}\\
&\mathbf{c}_t=\mathbf{f}_t\odot\mathbf{c}_{t-1}+\mathbf{i}_t\odot\tilde{\mathbf{c}}_t\tag{$10$}\label{eq:10}\\
&\mathbf{h}_t=\mathbf{o}_t\odot\tanh(\mathbf{c}_t)\tag{$11$}\label{eq:11}
\end{align*}
where $\mathbf{x}_t$ is the current input, $\sigma$ denotes logistic sigmoid function and $\odot$ denotes element-wise multiplication. By selectively controlling portions of the memory cell $\mathbf{c}_t$ to update, erase and forget at each time step, LSTM can better comprehend long-term dependencies with respect to labels of the whole sequences.

\subsection{A Generalized Architecture}

\begin{figure}[!ht]
\centering
\begin{subfigure}[b]{0.48\textwidth}
\centering
\includegraphics[width=\textwidth]{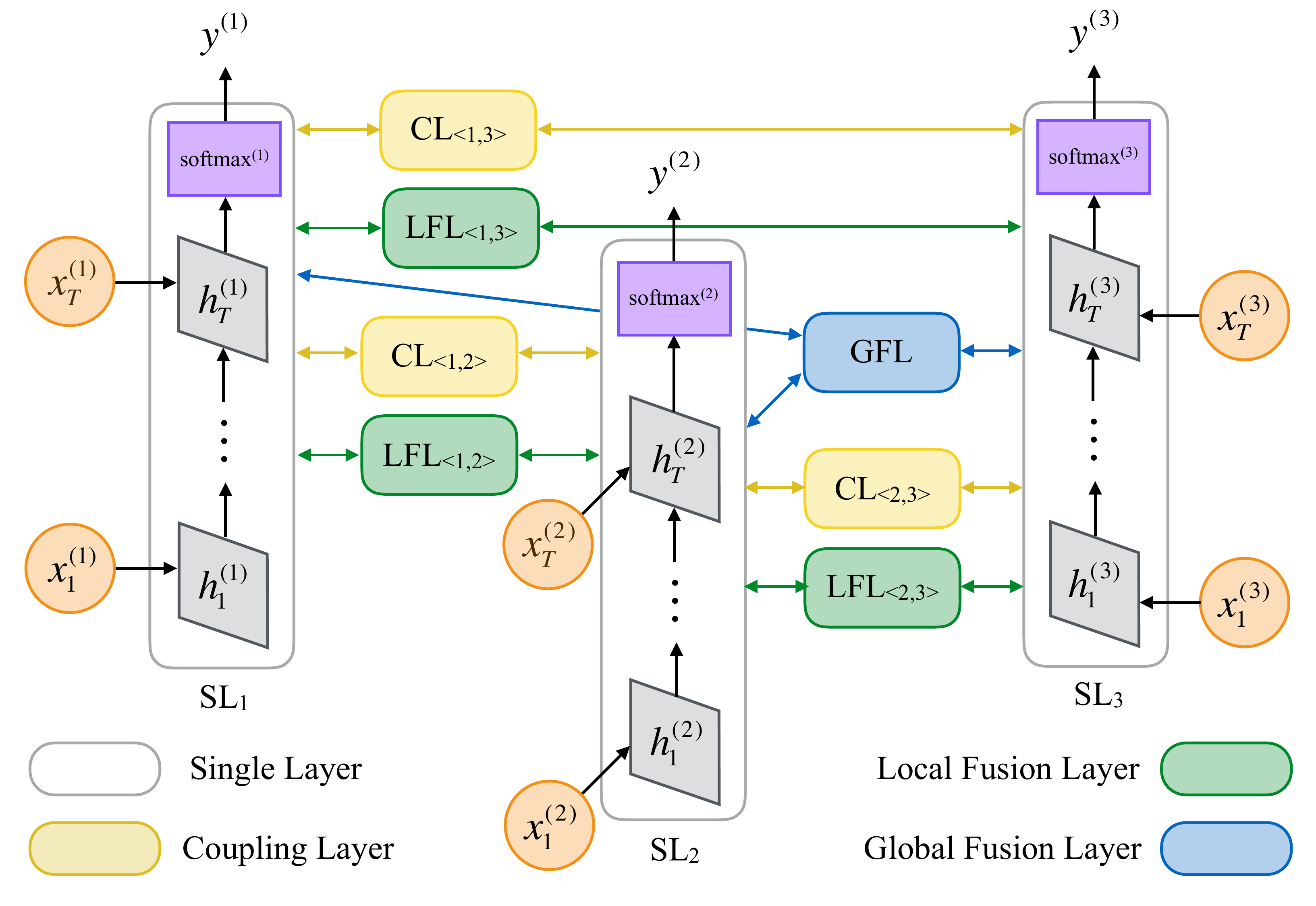}
\caption{Overall architecture with Single Layers, Coupling Layers, Local Fusion Layers and Global Fusion Layer}\label{architecture}
\end{subfigure}
\begin{subfigure}[b]{0.45\textwidth}
\centering
\includegraphics[width=\textwidth]{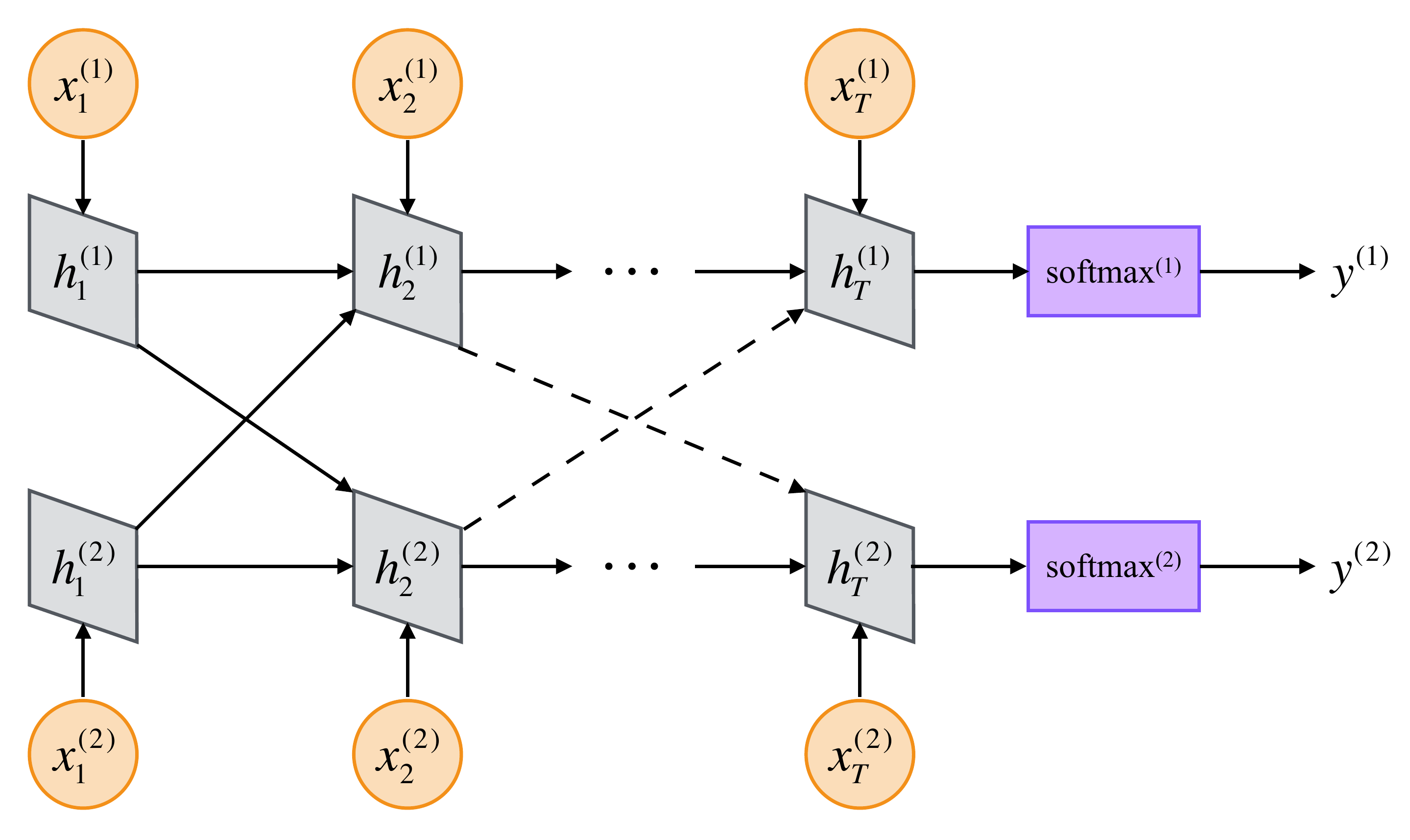}
\caption{Details of Coupling Layer Between $T_1$ and $T_2$}\label{coupling}
\end{subfigure}
\begin{subfigure}[b]{0.45\textwidth}
\centering
\includegraphics[width=\textwidth]{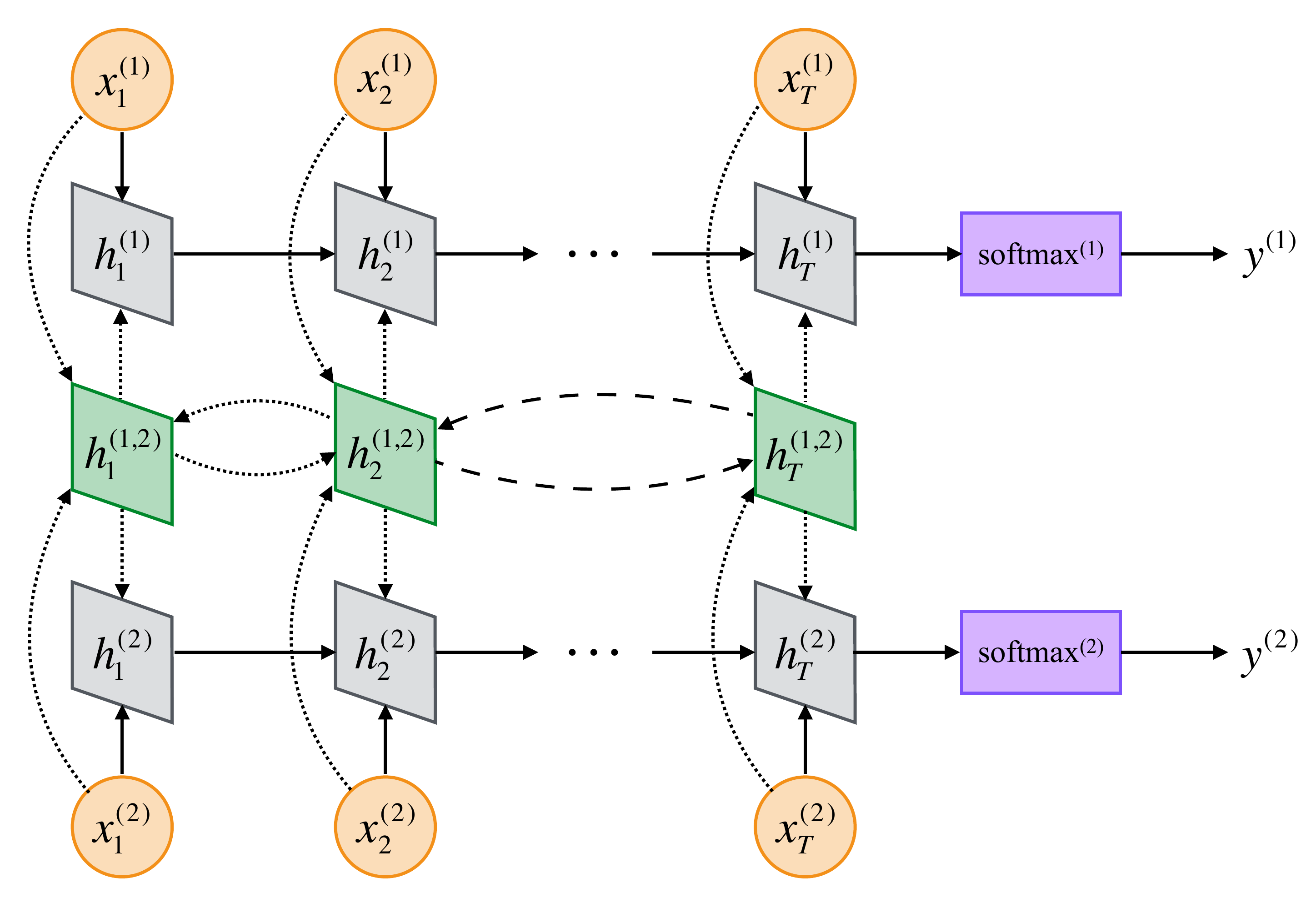}
\caption{Details of Local Fusion Layer Between $T_1$ and $T_2$}\label{fusion}
\end{subfigure}
\caption{A generalized recurrent neural architecture for modeling text with multi-task learning}\label{Layers}
\end{figure}

Based on the LSTM implementation of ~\cite{DBLP:journals/corr/Graves13}, we propose a generalized multi-task learning architecture for text classification with four types of recurrent neural layers to convey information inside and among tasks. Figure \ref{Layers} illustrates the structure design and information flows of our model, where three tasks are jointly learned in parallel.

As Figure \ref{architecture} shows, each task owns a LSTM-based \textbf{Single Layer} for intra-task learning. Pair-wise \textbf{Coupling Layer} and \textbf{Local Fusion Layer} are designed for direct and indirect inter-task interactions. And we further utilize a \textbf{Global Fusion Layer} to maintain a global memory for information shared among all tasks.

\subsubsection{Single Layer}

Each task owns a LSTM-based Single Layer with a collection of parameters $\Phi^{(k)}$, taking Eqs.(\ref{eq:9}) for example.
\begin{equation}\tag{$12$}\label{eq:12}
\tilde{\mathbf{c}}_t^{(k)}=\tanh(\mathbf{W}_c^{(k)}\mathbf{x}_t^{(k)}+\mathbf{U}_c^{(k)}\mathbf{h}_{t-1}^{(k)})
\end{equation}

Input sequences of each task are transformed into vector representations $(\mathbf{x}^{(1)}, \mathbf{x}^{(2)},...,\mathbf{x}^{(K)})$, which are later recurrently fed into the corresponding Single Layers. The hidden states at the last time step $\mathbf{h}_T^{(k)}$ of each Single Layer can be regarded as fix-length representations of the whole sequences, which are followed by a fully connected layer and a softmax non-linear layer to produce class distributions.
\begin{equation}\tag{$13$}\label{eq:13}
\hat{\mathbf{y}}^{(k)}=softmax(\mathbf{W}^{(k)}\mathbf{h}_T^{(k)}+\mathbf{b}^{(k)})
\end{equation}
where $\hat{\mathbf{y}}^{(k)}$ is the predicted class distribution for $\mathbf{x}^{(k)}$.

\subsubsection{Coupling Layer}

Besides Single Layers, we design Coupling Layers to model \textbf{direct} pair-wise interactions between tasks. For each pair of tasks, hidden states and memory cells of the Single Layers can obtain extra information directly from each other, as shown in Figure \ref{coupling}.

We re-define Eqs.(\ref{eq:12}) and utilize a gating mechanism to control the portion of information flows from one task to another. The memory content $\tilde{\mathbf{c}}_t^{(k)}$ of each Single Layer is updated on the leverage of pair-wise couplings.
\begin{align*}
&\tilde{\mathbf{c}}_t^{(k)}=\tanh(\mathbf{W}_c^{(k)}\mathbf{x}_t^{(k)}+\sum_{j=1}^K\mathbf{g}^{(j\rightarrow k)}\mathbf{U}_c^{(j\rightarrow k)}\mathbf{h}_{t-1}^{(j)})\tag{$14$}\label{eq:14}\\
&\mathbf{g}^{(j\rightarrow k)}=\sigma(\mathbf{W}_{gc}^{(k)}\mathbf{x}_t^{(k)}+\mathbf{U}_{gc}^{(j)}\mathbf{h}_{t-1}^{(j)})\tag{$15$}\label{eq:15}
\end{align*}
where $\mathbf{g}^{(j\rightarrow k)}$ controls the portion of information flow from $T_j$ to $T_k$, based on the correlation strength between $\mathbf{x}_t^{(k)}$ and $\mathbf{h}_{t-1}^{(j)}$ at the current time step.

In this way, the hidden states and memory cells of each Single Layer can obtain extra information from other tasks and stronger relevance results in higher chances of reception.

\subsubsection{Local Fusion Layer}

Different from Coupling Layers, Local Fusion Layers introduce a shared bi-directional LSTM Layer to model \textbf{indirect} pair-wise interactions between tasks. For each pair of tasks, we feed the Local Fusion Layer with the concatenation of both inputs, $\mathbf{x}_t^{(j,k)}=\mathbf{x}_t^{(j)}\oplus\mathbf{x}_t^{(k)}$, as shown in Figure \ref{fusion}. We denote the output of the Local Fusion Layer as $\mathbf{h}_t^{(j,k)}=\overrightarrow{\mathbf{h}}_t^{(j,k)}\oplus\overleftarrow{\mathbf{h}}_t^{(j,k)}$, a concatenation of hidden states from the forward and backward LSTM at each time step.

Similar to Coupling Layers, hidden states and memory cells of the Single Layers can selectively decide how much information to accept from the pair-wise Local Fusion Layers. We re-define Eqs.(\ref{eq:14}) by considering the interactions between the memory content $\tilde{\mathbf{c}}_t^{(k)}$ and outputs of the Local Fusion Layers as follows.
\begin{align*}
&\tilde{\mathbf{c}}_t^{(k)}=\tanh(\mathbf{W}_c^{(k)}\mathbf{x}_t^{(k)}+\mathbf{C}_t^{(k)}+\mathbf{LF}_t^{(k)}\tag{$16$})\label{eq:16}\\
&\mathbf{LF}_t^{(k)}=\sum_{j=1,j\neq k}^K\mathbf{g}^{(j,k)}\mathbf{U}_c^{(j,k)}\mathbf{h}_{t}^{(j,k)}\tag{$17$}\label{eq:17}\\
&\mathbf{g}^{(j,k)}=\sigma(\mathbf{W}_{gf}^{(k)}\mathbf{x}_t^{(k)}+\mathbf{U}_{gf}^{(j)}\mathbf{h}_{t}^{(j,k)})\tag{$18$}\label{eq:18}
\end{align*}
where $\mathbf{C}_t^{(k)}$ denotes the coupling term in Eqs.(\ref{eq:14}) and $\mathbf{LF}_t^{(k)}$ represents the local fusion term. Again, we employ a gating mechanism $\mathbf{g}^{(j,k)}$ to control the portion of information flow from the Local Coupling Layers to $T_k$.

\subsubsection{Global Fusion Layer}

Indirect interactions between Single Layers can be \textbf{pair-wise} or \textbf{global}, so we further propose the Global Fusion Layer as a shared memory storage among all tasks. The Global Fusion Layer consists of a bi-directional LSTM Layer with the inputs $\mathbf{x}_t^{(g)}=\mathbf{x}_t^{(1)}\oplus\mathbf{x}_t^{(2)}\oplus\cdots\oplus\mathbf{x}_t^{(K)}$ and the outputs $\mathbf{h}_t^{(g)}=\overrightarrow{\mathbf{h}}_t^{(g)}\oplus\overleftarrow{\mathbf{h}}_t^{(g)}$.

We denote the global fusion term as $\mathbf{GF}_t^{(k)}$ and the memory content $\tilde{\mathbf{c}}_t^{(k)}$ is calculated as follows.

\begin{align*}
&\tilde{\mathbf{c}}_t^{(k)}=\tanh(\mathbf{W}_c^{(k)}\mathbf{x}_t^{(k)}+\mathbf{C}_t^{(k)}+\mathbf{LF}_t^{(k)}+\mathbf{GF}_t^{(k)}\tag{$19$})\label{eq:19}\\
&\mathbf{GF}_t^{(k)}=\sigma(\mathbf{W}_{gg}^{(k)}\mathbf{x}_t^{(k)}+\mathbf{U}_{gg}^{(k)}\mathbf{h}_{t}^{(g)})\mathbf{U}_c^{(g)}\mathbf{h}_{t}^{(g)}\tag{$20$}\label{eq:20}
\end{align*}

As a result, our architecture covers complicated interactions among different tasks. It is capable of mapping a collection of input sequences from different tasks into a combination of predicted class distributions in parallel, as shown in Eqs.(\ref{eq:3}).

\subsection{Sampling \& Training}

Most previous multi-task learning models~\cite{DBLP:conf/icml/CollobertW08,DBLP:conf/naacl/LiuGHDDW15,DBLP:conf/emnlp/LiuQH16,DBLP:conf/ijcai/LiuQH16} belongs to \textbf{Type-\uppercase\expandafter{\romannumeral1}} or \textbf{Type-\uppercase\expandafter{\romannumeral2}}. The total number of input samples is $N=\sum_{k=1}^{K}N_k$, where $N_k$ are the sample numbers of each task.

However, our model focuses on \textbf{Type-\uppercase\expandafter{\romannumeral3}} and requires a 4-D tensor $N\times K\times T\times d$ as inputs, where $N,K,T,d$ are total number of input collections, task number, sequence length and embedding size respectively. Samples from different tasks are jointly learned in parallel so the total number of all possible input collections is $N_{max}=\prod_{k=1}^{K}N_k$. We propose a \textbf{Task Oriented Sampling} algorithm to generate sample collections for improvements of a specific task $T_k$.

\renewcommand{\algorithmicrequire}{\textbf{Input:}}
\renewcommand{\algorithmicensure}{\textbf{Output:}}
\begin{algorithm}[ht]
\caption{\textbf{T}ask \textbf{O}riented \textbf{S}ampling}\label{alg:1}
\begin{algorithmic}[1]
\Require 
$N_i$ samples from each task $T_i$;
$k$, the oriented task index;
$n_0$, upsampling coefficient s.t. $N=n_0 N_k$
\Ensure sequence collections $\mathbf{X}$ and label combinations $\mathbf{Y}$

\For{each $i\in [1,K]$} 
	\State generate a set $S_i$ with $N$ samples for each task: 
	\If{$i=k$} 
	\State repeat each sample for $n_0$ times
	\ElsIf{$N_i\geq N$} 
	\State randomly select $N$ samples without replacements
	\Else 
	\State randomly select $N$ samples with replacements
	\EndIf
\EndFor
\For{each $j\in [1,N]$} 
	\State randomly select a sample from each $S_i$ without replacements
	\State combine their features and labels as $X_j$ and $Y_j$
\EndFor
\State merge all $X_j$ and $Y_j$ to produce the sequence collections $\mathbf{X}$ and label combinations $\mathbf{Y}$ 
\end{algorithmic} 
\end{algorithm}

Given the generated sequence collections $\mathbf{X}$ and label combinations $\mathbf{Y}$, the overall loss function can be calculated based on Eqs.(\ref{eq:4}) and (\ref{eq:13}). The training process is conducted in a stochastic manner until convergence. For each loop, we randomly select a collection from the $N$ candidates and update the parameters by taking a gradient step.

\section{Experiment}

In this section, we design three different scenarios of multi-task learning based on five benchmark datasets for text classification. we investigate the empirical performances of our model and compare it to existing state-of-the-art models.

\begin{table*}[!ht]
\centering
\caption{Five benchmark classification datasets: SST, IMDB, MDSD, RN, QC.}\label{tab:1}
\begin{tabular}{|c|m{7cm}|c|c|c|c|c|} \hline
Dataset & Description & Type & Length & Class & Objective \\ \hline
SST & Movie reviews in Stanford Sentiment Treebank including SST-1 and SST-2 & Sentence & 19 / 19 & 5 / 2 & Sentiment \\ \hline
IMDB & Internet Movie Database & Document & 279 & 2 & Sentiment \\ \hline
MDSD & Product reviews on books, DVDs, electronics and kitchen appliances & Document & 176 / 189 / 115 / 97 & 2 & Sentiment \\ \hline
RN & Reuters Newswire topics classification & Document & 146 & 46 & Topics \\ \hline
QC & Question Classification & Sentence & 10 & 6 & Question Types \\ \hline
\end{tabular}
\end{table*}

\subsection{Datasets}

As Table \ref{tab:1} shows, we select five benchmark datasets for text classification and design three experiment scenarios to evaluate the performances of our model.

\begin{itemize}
\item \textbf{Multi-Cardinality} Movie review datasets with different average lengths and class numbers, including \textbf{SST-1}~\cite{Socher-etal:2013}, \textbf{SST-2} and \textbf{IMDB}~\cite{maas-EtAl:2011:ACL-HLT2011}. 
\item \textbf{Multi-Domain} Product review datasets on different domains from \textit{Multi-Domain Sentiment Dataset}~\cite{DBLP:conf/acl/BlitzerDP07}, including \textbf{Books}, \textbf{DVDs}, \textbf{Electronics} and \textbf{Kitchen}.
\item \textbf{Multi-Objective} Classification datasets with different objectives, including \textbf{IMDB}, \textbf{RN}~\cite{DBLP:journals/tois/ApteDW94} and \textbf{QC}~\cite{DBLP:conf/coling/LiR02}.
\end{itemize}

\subsection{Hyperparameters and Training}

The whole network is trained through back propagation with stochastic gradient descent~\cite{DBLP:journals/ijon/Amari93}. We obtain a pre-trained lookup table by applying \textit{Word2Vec}~\cite{DBLP:journals/corr/abs-1301-3781} on the Google News corpus, which contains more than 100B words with a vocabulary size of about 3M. All involved parameters are randomly initialized from a truncated normal distribution with zero mean and standard deviation.

For each task $T_k$, we conduct \textbf{TOS} with $n_0=2$ to improve its performance. After training our model on the generated sample collections, we evaluate the performance of task $T_k$ by comparing $\hat{\mathbf{y}}^{(k)}$ and $\mathbf{y}^{(k)}$ on the test set. We apply 10-fold cross-validation and different combinations of hyperparameters are investigated, of which the best one, as shown in Table \ref{tab:2}, is reserved for comparisons with state-of-the-art models.

\begin{table}[!ht]
\centering
\caption{Hyperparameter settings}\label{tab:2}
\begin{tabular}{|c|c|} \hline
Embedding size & $d=300$ \\ \hline
Hidden layer size of LSTM & $n=100$ \\ \hline
Initial learning rate & $\eta=0.1$ \\ \hline
Regularization weight & $\lambda=10^{-5}$ \\ \hline
\end{tabular}
\end{table}

\subsection{Results}

We compare performances of our model with the implementation of ~\cite{DBLP:journals/corr/Graves13} and the results are shown in Table \ref{tab:3}. Our model obtains better performances in Multi-Domain scenario with an average improvement of 4.5\%, where datasets are product reviews on different domains with similar sequence lengths and the same class number, thus producing stronger correlations. Multi-Cardinality scenario also achieves significant improvements of 2.77\% on average, where datasets are movie reviews with different cardinalities. 

However, Multi-Objective scenario benefits less from multi-task learning due to lacks of salient correlation among sentiment, topic and question type. The QC dataset aims to classify each question into six categories and its performance even gets worse, which may be caused by potential noises introduced by other tasks. In practice, the structure of our model is flexible, as couplings and fusions between some empirically unrelated tasks can be removed to alleviate computation costs.

\begin{table*}[!ht]
\centering
\caption{Results of our model on different scenarios}\label{tab:3}
\begin{tabular}{|c|c|c|c|c|c|c|c|c|c|c|}
\hline
\multirow{2}{*}{\textbf{Model}} & \multicolumn{3}{|c|}{\textbf{Multi-Cardinality}} & \multicolumn{4}{|c|}{\textbf{Multi-Domain}} & \multicolumn{3}{|c|}{\textbf{Multi-Objective}} \\ \cline{2-11} 
& SST-1 & SST-2 & IMDB & Books & DVDs & Electronics & Kitchen & IMDB & RN & QC \\ \hline
Single Task & 45.9 & 85.8 & 88.5 & 78.0 & 79.5 & 81.2 & 81.8 & 88.5 & 83.6 & \textbf{92.5} \\ \hline
Our Model & \textbf{49.2} & \textbf{87.7} & \textbf{91.6} & \textbf{83.5} & \textbf{84.0} & \textbf{86.2} & \textbf{84.8} & \textbf{89.7} & \textbf{84.2} & 92.3 \\ \hline
\end{tabular}
\end{table*}

\subsubsection{Influences of $n_0$ in TOS}

We further explore the influence of $n_0$ in TOS on our model, which can be any positive integer. A higher value means larger and more various samples combinations, while requires higher computation costs.

\begin{figure}[!ht]
\centering
\includegraphics[width=0.4\textwidth]{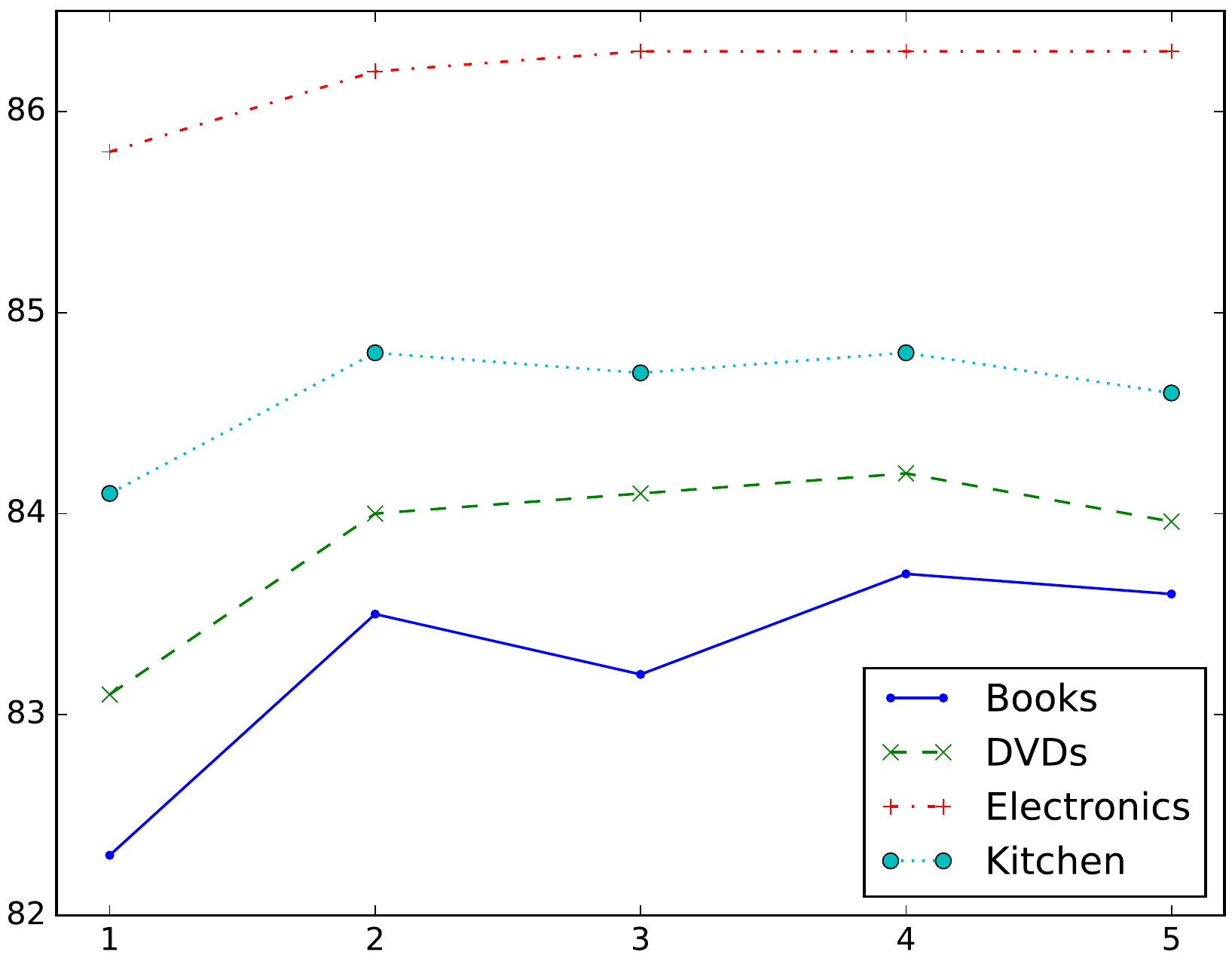}
\caption{Influences of $n_0$ in TOS on different datasets}\label{n0}
\end{figure}

Figure \ref{n0} shows the performances of datasets in Multi-Domain scenario with different $n_0$. Compared to $n_0=1$, our model can achieve considerable improvements when $n_0=2$ as more samples combinations are available. However, there are no more salient gains as $n_0$ gets larger and potential noises from other tasks may lead to performance degradations. For a trade-off between efficiency and effectiveness, we determine $n_0=2$ as the optimal value for our experiments.

\subsubsection{Pair-wise Performance Gain}

In order to measure the correlation strength between two task $T_i$ and $T_j$, we learn them jointly with our model and define Pair-wise Performance Gain as $PPG_{ij}=\sqrt{\frac{{P_i}'{P_j}'}{P_iP_j}}$, where $P_i,P_j,{P_i}',{P_j}'$ are the performances of tasks $T_i$ and $T_j$ when learned individually and jointly.

% \begin{equation}
% PPG_{ij}=\sqrt{\frac{{P_i}'{P_j}'}{P_iP_j}}\tag{$21$}\label{eq:21}
% \end{equation}

\begin{figure}[!ht]
\centering
\includegraphics[width=0.4\textwidth]{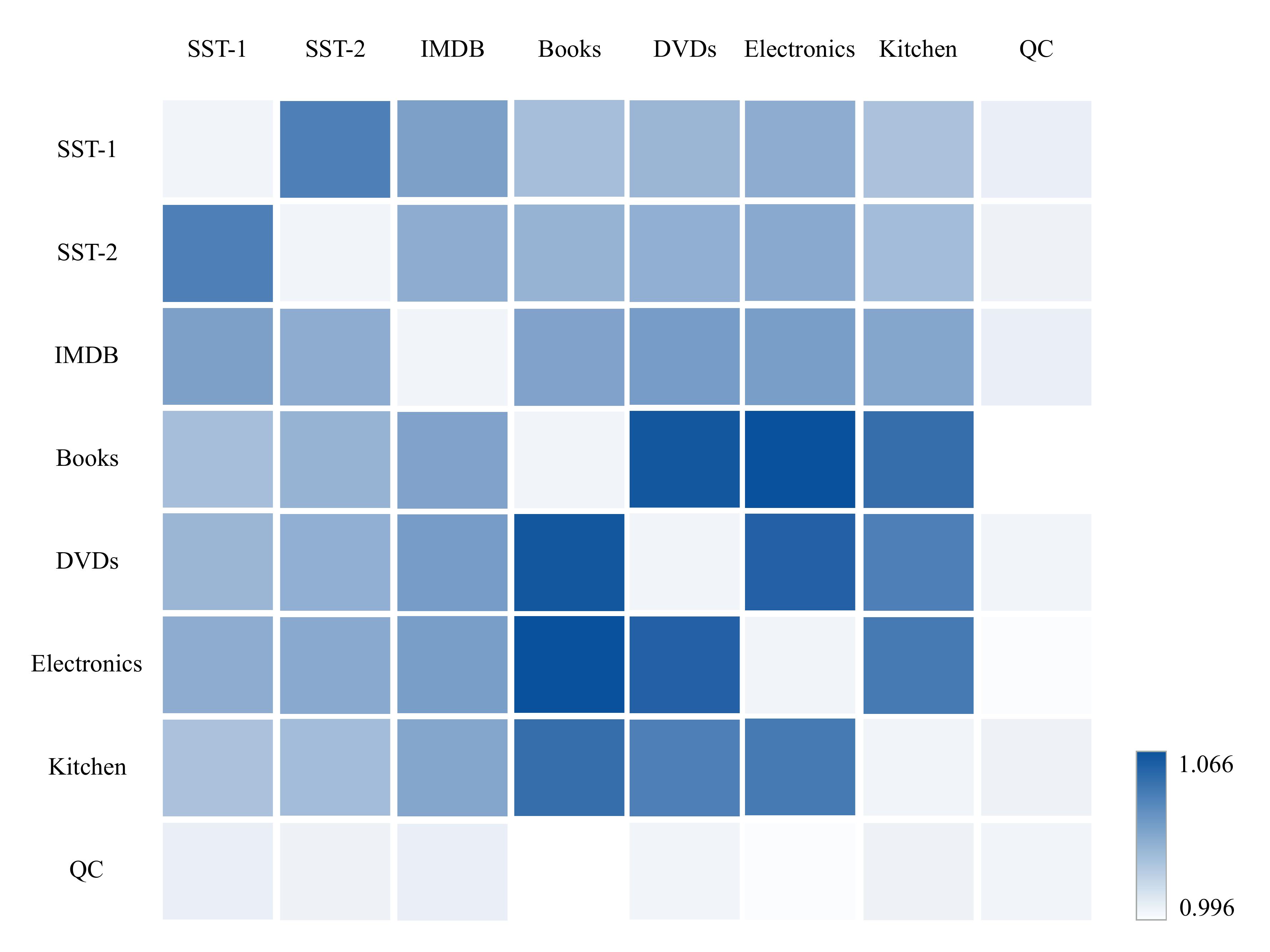}
\caption{Visualization of Pair-wise Performance Gains}\label{ppg}
\end{figure}

We calculate PPGs for every two tasks in Table \ref{tab:1} and illustrate the results in Figure \ref{ppg}, where darkness of colors indicate strength of correlation. It is intuitive that datasets of Multi-Domain scenario obtain relatively higher PPGs with each other as they share similar cardinalities and abundant low-level linguistic characteristics. Sentences of QC dataset are much shorter and convey unique characteristics from other tasks, thus resulting in quite lower PPGs.

\begin{table*}[!ht]
\centering
\caption{Comparisons with state-of-the-art models}\label{tab:4}
\begin{tabular}{|c|c|c|c|c|c|c|c|c|c|c|}
\hline
\textbf{Model} & SST-1 & SST-2 & IMDB & Books & DVDs & Electronics & Kitchen & QC \\ \hline
NBOW & 42.4 & 80.5 & 83.62 & - & - & - & - & 88.2 \\ \hline
% DCNN & 48.5 & 86.8 & - & - & - & - & - & 93.0 \\ \hline
PV & 44.6 & 82.7 & \textbf{91.7} & - & - & - & - & 91.8 \\ \hline
MT-RNN & \textbf{49.6} & \textbf{87.9} & 91.3 & - & - & - & - & - \\ \hline
MT-CNN & - & - & - & 80.2 & 81.0 & 83.4 & 83.0 & - \\ \hline
MT-DNN & - & - & - & 79.7 & 80.5 & 82.5 & 82.8 & - \\ \hline
% DSM & 49.5 & 87.8 & 91.2 & 82.8 & 83.0 & 85.5 & 84.0 & - \\ \hline
GRNN & 47.5 & 85.5 & - & - & - & - & - & \textbf{93.8} \\ \hline
\textbf{Our Model} & 49.2 & 87.7 & 91.6 & \textbf{83.5} & \textbf{84.0} & \textbf{86.2} & \textbf{84.8} & 92.3 \\ \hline
\end{tabular}
\end{table*}

\subsection{Comparisons with State-of-the-art Models}

We apply the optimal hyperparameter settings and compare our model against the following state-of-the-art models:

\begin{itemize}
\item \textbf{NBOW} Neural Bag-of-Words that simply sums up embedding vectors of all words.
% \item \textbf{DCNN} Dynamic Convolutional Neural Network with dynamic k-max pooling~\cite{DBLP:conf/acl/KalchbrennerGB14}.
\item \textbf{PV} Paragraph Vectors followed by logistic regression~\cite{DBLP:conf/icml/LeM14}.
\item \textbf{MT-RNN} Multi-Task learning with Recurrent Neural Networks by a shared-layer architecture~\cite{DBLP:conf/ijcai/LiuQH16}. 
\item \textbf{MT-CNN} Multi-Task learning with Convolutional Neural Networks~\cite{DBLP:conf/icml/CollobertW08} where lookup tables are partially shared. 
\item \textbf{MT-DNN} Multi-Task learning with Deep Neural Networks~\cite{DBLP:conf/naacl/LiuGHDDW15} that utilizes bag-of-word representations and a hidden shared layer.
% \item \textbf{DSM} Deep multi-task learning with Shared Memory~\cite{DBLP:conf/emnlp/LiuQH16} where a external memory and a reading/writing mechanism are introduced.
\item \textbf{GRNN} Gated Recursive Neural Network for sentence modeling~\cite{DBLP:conf/emnlp/ChenQZWH15}.
\end{itemize}

As Table \ref{tab:4} shows, our model obtains competitive or better performances on all tasks except for the QC dataset, as it contains poor correlations with other tasks. MT-RNN slightly outperforms our model on SST, as sentences from this dataset are much shorter than those from IMDB and MDSD, and another possible reason may be that our model are more complex and requires larger data for training. Our model proposes the designs of various interactions including coupling, local and global fusion, which can be further implemented by other state-of-the-art models and produce better performances.

\section{Related Work}

There are a large body of literatures related to multi-task learning with neural networks in NLP~\cite{DBLP:conf/icml/CollobertW08,DBLP:conf/naacl/LiuGHDDW15,DBLP:conf/emnlp/LiuQH16,DBLP:conf/ijcai/LiuQH16}.

~\cite{DBLP:conf/icml/CollobertW08} belongs to \textbf{Type-\uppercase\expandafter{\romannumeral1}} and utilizes shared lookup tables for common features, followed by task-specific neural layers for several traditional NLP tasks such as part-of-speech tagging and semantic parsing. They use a fix-size window to solve the problem of variable-length texts, which can be better handled by recurrent neural networks.

~\cite{DBLP:conf/naacl/LiuGHDDW15,DBLP:conf/emnlp/LiuQH16,DBLP:conf/ijcai/LiuQH16} all belong to \textbf{Type-\uppercase\expandafter{\romannumeral2}} where samples from different tasks are learned sequentially. ~\cite{DBLP:conf/naacl/LiuGHDDW15} applies bag-of-word representation and information of word orders are lost. ~\cite{DBLP:conf/emnlp/LiuQH16} introduces an external memory for information sharing with a reading/writing mechanism for communicating, and ~\cite{DBLP:conf/ijcai/LiuQH16} proposes three different models for multi-task learning with recurrent neural networks. However, models of these two papers only involve pair-wise interactions, which can be regarded as specific implementations of Coupling Layer and Fusion Layer in our model.

Different from the above models, our model focuses on \textbf{Type-\uppercase\expandafter{\romannumeral3}} and utilize recurrent neural networks to comprehensively capture various interactions among tasks, both direct and indirect, local and global. Three or more tasks are learned simultaneously and samples from different tasks are trained in parallel benefitting from each other, thus obtaining better sentence representations.

\section{Conclusion and Future Work}

In this paper, we propose a multi-task learning architecture for text classification with four types of recurrent neural layers. The architecture is structurally flexible and can be regarded as a generalized case of many previous works with deliberate designs. We explore three different scenarios of multi-task learning and our model can improve performances of most tasks with additional related information from others in all scenarios.

In future work, we would like to investigate further implementations of couplings and fusions, and conclude more multi-task learning perspectives. 

\bibliographystyle{named}
\bibliography{ijcai17}

\end{document}